\documentclass{article}
\usepackage{ijcai16}
\usepackage{times}
\usepackage[T1]{fontenc}
\usepackage{graphicx}
\usepackage{caption}
\usepackage{subcaption}
\usepackage[english]{babel}
\usepackage{arydshln}
\usepackage[utf8]{inputenc}
\usepackage{CJK}
\usepackage{multirow}
\usepackage{url}

\newcommand{\textcite}[1]{\citeauthor{#1} [\citeyear{#1}]}

\title{Exploring Segment Representations for Neural Segmentation Models}
\author{Yijia Liu, Wanxiang Che \thanks{Email corresponding.}, Jiang Guo, Bing Qin, Ting Liu \\
  Research Center for Social Computing and Information Retrieval \\
  Harbin Institute of Technology, China \\
  {\tt \{yjliu,car,jguo,qinb,tliu\}@ir.hit.edu.cn} \\}

\begin{document}

\maketitle

\begin{abstract}
Many natural language processing (NLP) tasks can be generalized into segmentation problem.
In this paper, we combine semi-CRF with neural network to solve NLP segmentation tasks.
Our model represents a segment both by composing the input units and embedding the entire segment.
We thoroughly study different composition functions and different segment embeddings.
We conduct extensive experiments on two typical segmentation tasks: named entity recognition (NER) and Chinese word segmentation (CWS).
Experimental results show that our neural semi-CRF model benefits from representing the entire segment and achieves the state-of-the-art performance on CWS benchmark dataset and competitive results on the CoNLL03 dataset.
\end{abstract}

\section{Introduction}

Given an input sequence, {\it segmentation} is the problem of identifying and assigning tags to its subsequences.
Many natural language processing (NLP) tasks can be cast into the segmentation problem, like named entity recognition \cite{okanohara-EtAl:2006:COLACL}, opinion extraction \cite{yang-cardie:2012:EMNLP-CoNLL}, and Chinese word segmentation \cite{andrew:2006:EMNLP}.
Properly representing {\it segment} is critical for good segmentation performance.
Widely used sequence labeling models like conditional random fields \cite{Lafferty:2001:CRF:645530.655813} represent the contextual information of the segment boundary as a proxy to entire segment and achieve segmentation by labeling input units (e.g. words or characters) with boundary tags.
Compared with sequence labeling model, models that directly represent segment are attractive because they are not bounded by local tag dependencies and can effectively adopt segment-level information.
Semi-Markov CRF (or semi-CRF) \cite{NIPS2005_427} is one of the models that directly represent the entire segment.
In semi-CRF, the conditional probability of a semi-Markov chain on the input sequence is explicitly modeled, whose each state corresponds to a subsequence of input units, which makes semi-CRF a natural choice for segmentation problem.

However, to achieve good segmentation performance, conventional semi-CRF models require carefully hand-crafted features to represent the segment.
Recent years witness a trend of applying neural network models to NLP tasks.
The key strengths of neural approaches in NLP are their ability for modeling the compositionality of language and learning distributed representation from large-scale unlabeled data.
Representing a segment with neural network is appealing in semi-CRF because various neural network structures \cite{Hochreiter:1997:LSM:1246443.1246450} have been proposed to compose sequential inputs of a segment and the well-studied word embedding methods \cite{DBLP:journals/corr/MikolovSCCD13} make it possible to learn entire segment representation from unlabeled data.

In this paper, we combine neural network with semi-CRF and make a thorough study on the problem of representing a segment in neural semi-CRF.
\textcite{DBLP:journals/corr/KongDS15} proposed a segmental recurrent neural network (SRNN) which represents a segment by composing input units with RNN.
We study alternative network structures besides the SRNN.
We also study segment-level representation using {\it segment embedding} which encodes the entire segment explicitly.
We conduct extensive experiments on two typical NLP segmentation tasks: named entity recognition (NER) and Chinese word segmentation (CWS).
Experimental results show that our concatenation alternative achieves comparable performance with the original SRNN but runs 1.7 times faster and our neural semi-CRF greatly benefits from the segment embeddings.
In the NER experiments, our neural semi-CRF model with segment embeddings achieves an improvement of 0.7 F-score over the baseline and the result is competitive with state-of-the-art systems.
In the CWS experiments, our model achieves more than 2.0 F-score improvements on average.
On the PKU and MSR datasets, state-of-the-art F-scores of 95.67\% and 97.58\% are achieved respectively.
We release our code at \url{https://github.com/ExpResults/segrep-for-nn-semicrf}.

\section{Problem Definition}

\begin{figure}[t]
\centering
\includegraphics[width=0.9\columnwidth,trim={0cm 0.4cm 7.8cm 12cm},clip]{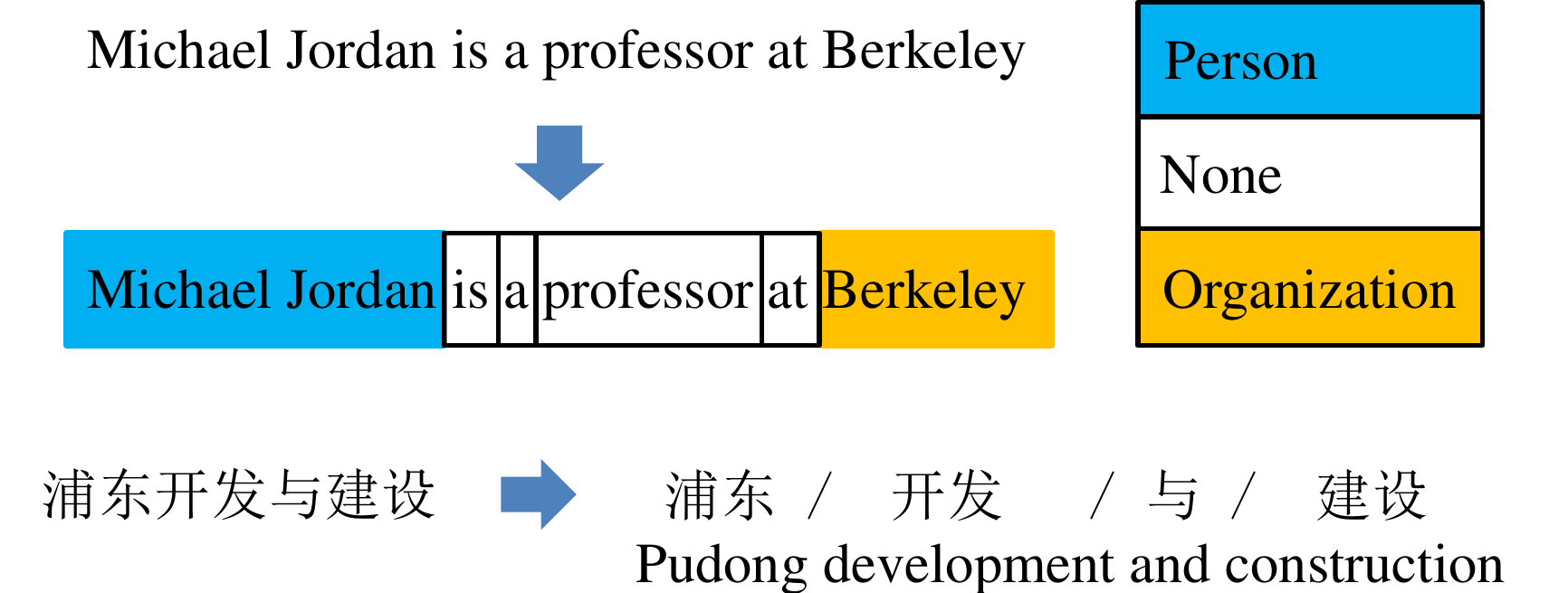}
\caption{Examples for named entity recognition (above) and Chinese word segmentation (below).}
\label{fig:ne-and-cws}
\end{figure}

Figure \ref{fig:ne-and-cws} shows examples of named entity recognition and Chinese word segmentation.
For the input word sequence in the NER example, its segments ({\it ``Michael Jordan'':PER, ``is'':NONE, ``a'':NONE, ``professor'':NONE, ``at'':NONE, ``Berkeley'':ORG}) reveal that ``Michaels Jordan'' is a person name and ``Berkeley'' is an organization.
In the CWS example, the subsequences (\begin{CJK*}{UTF8}{gkai}``浦东/Pudong'', ``开发/development'', ``与/and'', ``建设/construction''\end{CJK*}) of the input character sequence are recognized as words.
Both NER and CWS take an input sequence and partition it into disjoint subsequences.

Formally, for an input sequence $\mathbf{x}=( x_1, .., x_{|\mathbf{x}|} )$ of length $|\mathbf{x}|$, let $x_{a:b}$ denote its subsequence $(x_a,...,x_b)$.
A {\it segment} of $\mathbf{x}$ is defined as $(u, v, y)$ which means the subsequence $x_{u:v}$ is associated with label $y$.
A {\it segmentation} of $\mathbf{x}$ is a {\it segment} sequence $\mathbf{s} = (s_1,..,s_p)$, where $s_j=(u_j,v_j,y_j)$ and $u_{j+1}=v_j+1$.
Given an input sequence $\mathbf{x}$, the {\it segmentation} problem can be defined as the problem of finding $\mathbf{x}$'s most probable {\it segment} sequence $\mathbf{s}$.

\begin{figure*}[t]
\begin{subfigure}[b]{0.241\textwidth}
\centering
\includegraphics[width=\columnwidth,trim={3cm 2.8cm 5cm 5cm},clip]{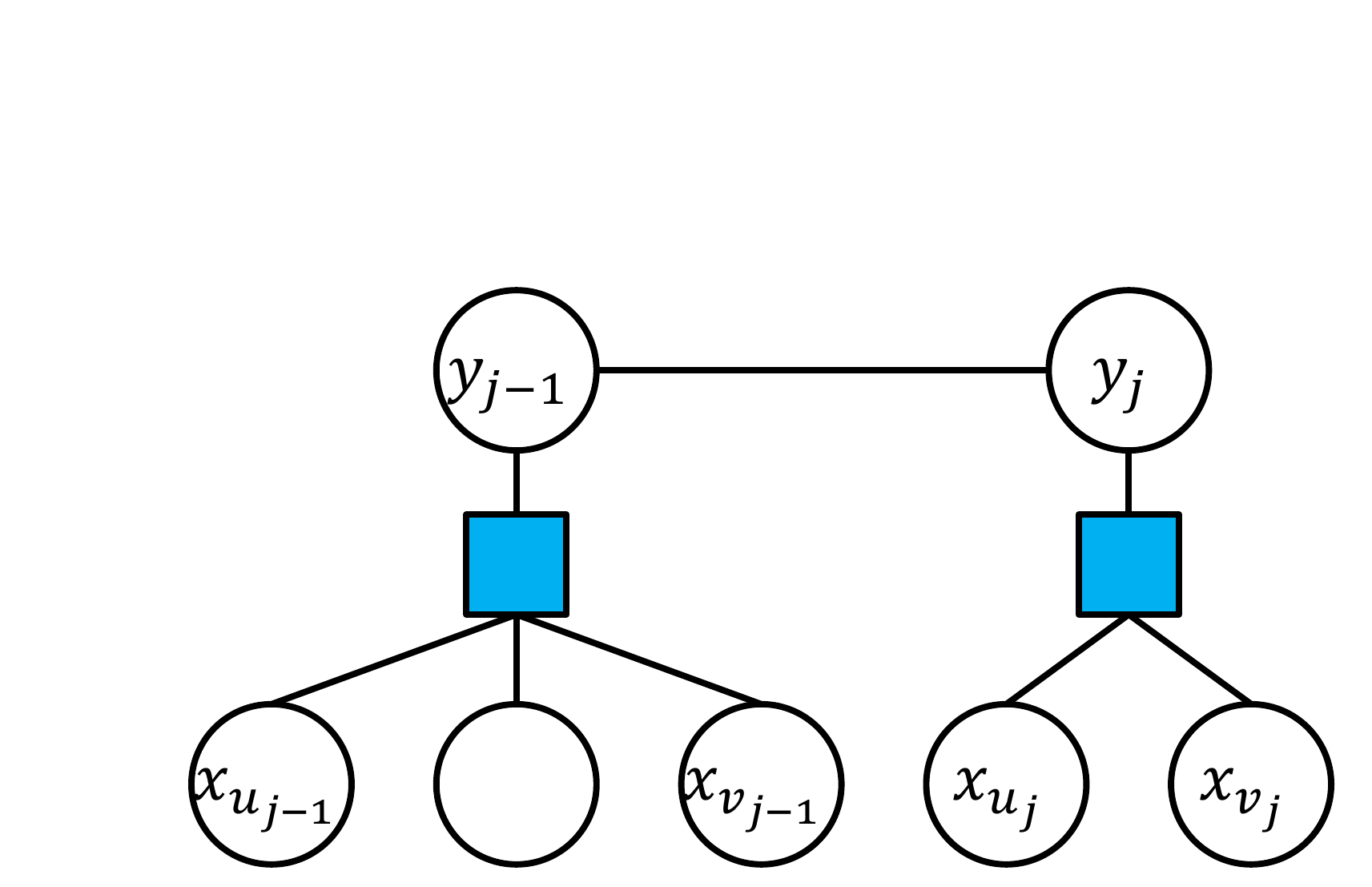}
\caption{semi-CRF}\label{fig:std}
\end{subfigure}
~
\begin{subfigure}[b]{0.241\textwidth}
\centering
\includegraphics[width=\columnwidth,trim={3cm 2.8cm 5cm 3.9cm},clip]{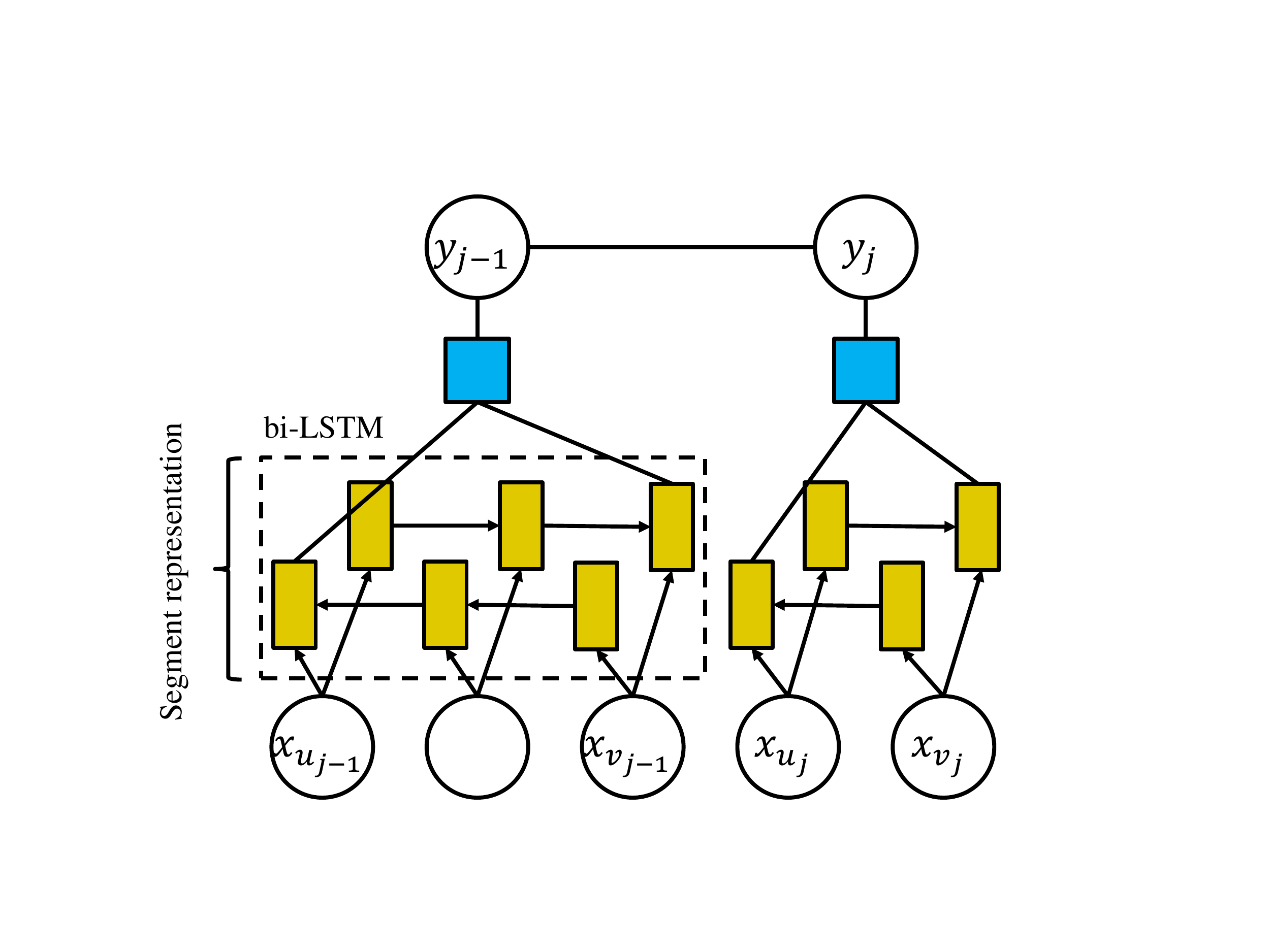}
\caption{SRNN}\label{fig:rnn}
\end{subfigure}
~
\begin{subfigure}[b]{0.241\textwidth}
\centering
\includegraphics[width=\columnwidth,trim={3cm 2.8cm 5cm 3.9cm},clip]{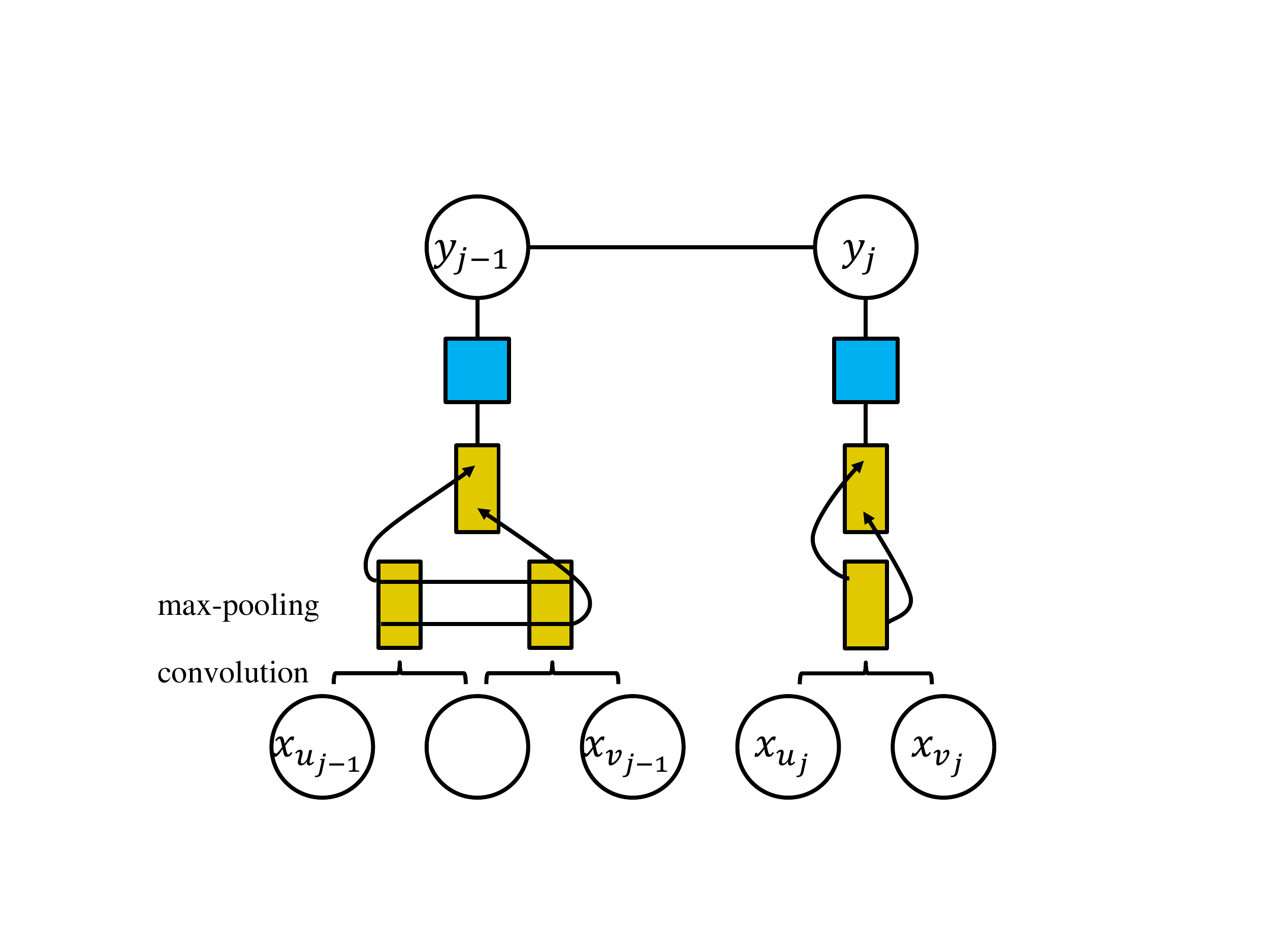}
\caption{SCNN}\label{fig:cnn}
\end{subfigure}
~
\begin{subfigure}[b]{0.241\textwidth}
\centering
\includegraphics[width=\columnwidth,trim={3cm 2.8cm 5cm 3.9cm},clip]{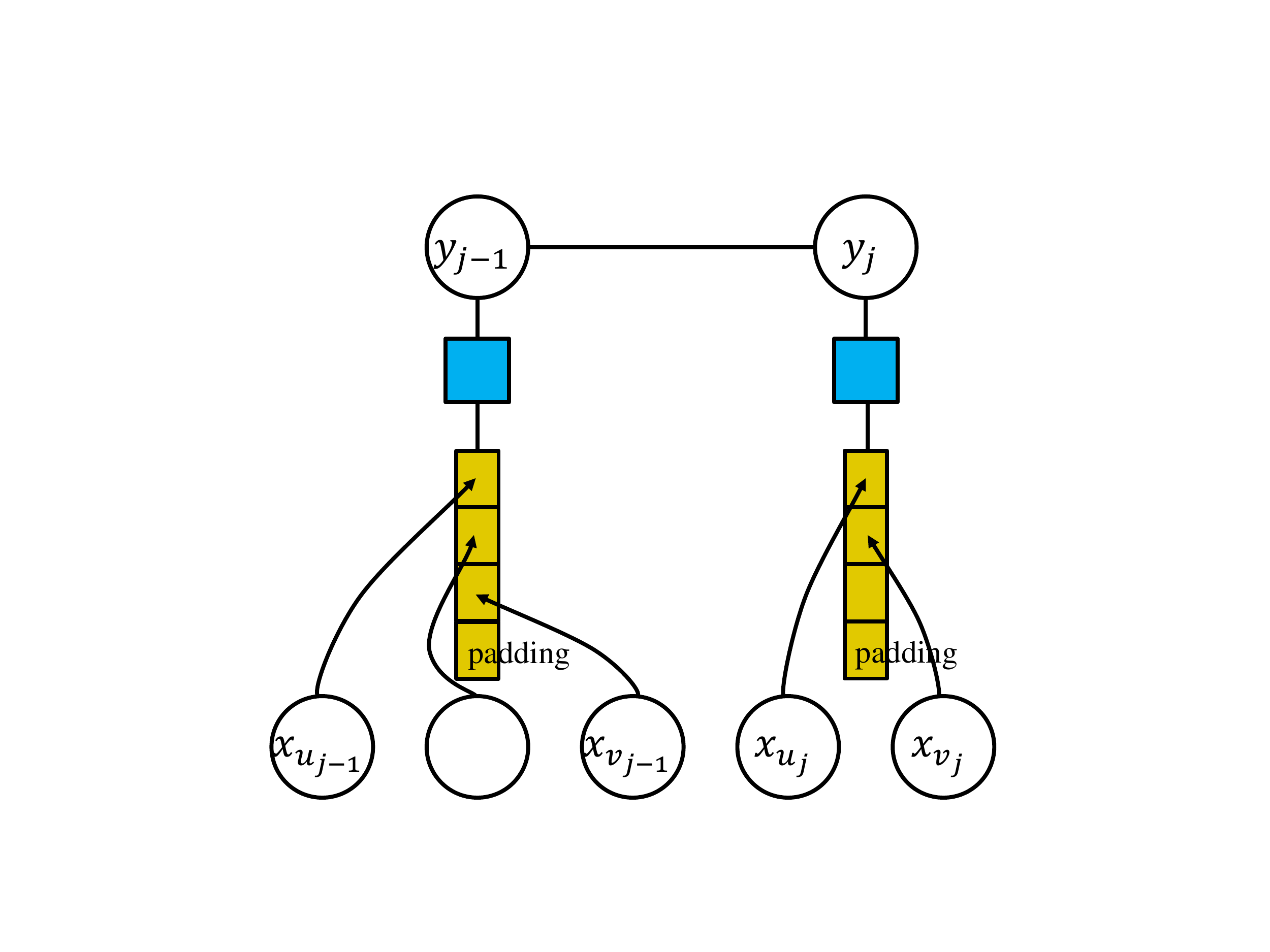}
\caption{SCONCATE}\label{fig:concate}
\end{subfigure}
\caption{An illustration for the semi-CRF, SRNN, SCNN and SCONCATE.
In these figures, circles represent the inputs, blue rectangles represent {\it factors} in graphic model and yellow rectangles represent generic nodes in the neural network model.}
\label{fig:semi-crf-vs-srnn}
\end{figure*}

\section{Neural Semi-Markov CRF}

Semi-Markov CRF (or semi-CRF, Figure \ref{fig:std}) \cite{NIPS2005_427} models the conditional probability of $\mathbf{s}$ on $\mathbf{x}$ as
\[
p(\mathbf{s}|\mathbf{x})=\frac{1}{Z(\mathbf{x})}\exp\{W \cdot G(\mathbf{x},\mathbf{s})\}
\]
where $G(\mathbf{x},\mathbf{s})$ is the feature function, $W$ is the weight vector and $Z(\mathbf{x})=\sum_{\mathbf{s'}\in \mathbf{S}} \exp\{W \cdot G(\mathbf{x}, \mathbf{s'})\}$ is the normalize factor of all possible {\it segmentations} $\mathbf{S}$ over $\mathbf{x}$.

By restricting the scope of feature function within a segment and ignoring label transition between segments (0-order semi-CRF), $G(\mathbf{x},\mathbf{s})$ can be decomposed as $G(\mathbf{x}, \mathbf{s})=\sum_{j=1}^p g(\mathbf{x},s_j)$ where $g(\mathbf{x},s_j)$ maps segment $s_j$ into its representation.
Such decomposition allows using efficient dynamic programming algorithm for inference.
To find the best segmentation in semi-CRF, let $\alpha_j$ denote the best segmentation ends with $j$\textsuperscript{th} input and $\alpha_j$ is recursively calculated as
\[\mathbf{\alpha}_j=\max_{l=1..L,y} \Psi(j-l,j,y) + \mathbf{\alpha}_{j-l-1}\]
where $L$ is the maximum length manually defined and $\Psi(j-l,j,y)$ is the transition weight for $s=(j-l,j,y)$ in which $\Psi(j-l,j,y)=W\cdot g(\mathbf{x}, s)$.

Previous semi-CRF works \cite{NIPS2005_427,okanohara-EtAl:2006:COLACL,andrew:2006:EMNLP,yang-cardie:2012:EMNLP-CoNLL} parameterize $g(\mathbf{x},s)$ as a sparse vector, each dimension of which represents the value of corresponding feature function.
Generally, these feature functions fall into two types: 1) the {\it CRF style features} which represent input unit-level information such as ``the specific words at location $i$''  2) the {\it semi-CRF style features} which represent segment-level information such as ``the length of the segment''.

\textcite{DBLP:journals/corr/KongDS15} proposed the segmental recurrent neural network model (SRNN, see Figure \ref{fig:rnn}) which combines the semi-CRF and the neural network model.
In SRNN, $g(\mathbf{x}, s)$ is parameterized as a bidirectional LSTM (bi-LSTM).
For a segment $s_j=(u_j,v_j,y_j)$, each input unit $x$ in subsequence $(x_{u_j},..,x_{v_j})$ is encoded as {\it embedding} and fed into the bi-LSTM.
The rectified linear combination of the final hidden layers from bi-LSTM is used as $g(\mathbf{x}, s)$.
\textcite{DBLP:journals/corr/KongDS15} pioneers in representing a segment in neural semi-CRF.
Bi-LSTM can be regarded as ``neuralized'' {\it CRF style features} which model the input unit-level compositionality.
However, in the SRNN work, only the bi-LSTM was employed without considering other input unit-level composition functions.
What is more, the {\it semi-CRF styled} segment-level information as an important representation was not studied.
In the following sections, we first study alternative input unit-level composition functions (\ref{sec:alt-inp-rep}).
Then, we study the problem of representing a segment at segment-level (\ref{sec:seg-rep}).

\subsection{Alternative Seg-Rep. via Input Composition}\label{sec:alt-inp-rep}

\subsubsection{Segmental CNN}
Besides recurrent neural network (RNN) and its variants, another widely used neural network architecture for composing and representing variable-length input is the convolutional neural network (CNN) \cite{Collobert:2011:NLP:1953048.2078186}.
In CNN, one or more filter functions are employed to convert a fix-width segment in sequence into one vector.
With filter function ``sliding'' over the input sequence, contextual information is encoded.
Finally, a pooling function is used to merge the vectors into one.
In this paper, we use a filter function of width 2 and max-pooling function to compose input units of a segment.
Following SRNN, we name our CNN segment representation as SCNN (see Figure \ref{fig:cnn}).

However, one problem of using CNN to compose input units into segment representation lies in the fact that the max-pooling function is insensitive to input position.
Two different segments sharing the same vocabulary can be treated without difference.
In a CWS example, \begin{CJK*}{UTF8}{gkai}``球拍卖'' (racket for sell) and ``拍卖球'' (ball audition)\end{CJK*} will be encoded into the same vector in SCNN if the vector of \begin{CJK*}{UTF8}{gkai}``拍卖''\end{CJK*} that produced by filter function is always preserved by max-pooling.

\subsubsection{Segmental Concatenation}
Concatenation is also widely used in neural network models to represent fixed-length input.
Although not designed to handle variable-length input, we see that in the inference of semi-CRF, a maximum length $L$ is adopted, which make it possible to use padding technique to transform the variable-length representation problem into fixed-length of $L$.
Meanwhile, concatenation preserves the positions of inputs because they are directly mapped into the certain positions in the resulting vector.
In this paper, we study an alternative concatenation function to compose input units into segment representation, namely the SCONCATE model (see Figure \ref{fig:concate}).
Compared with SRNN, SCONCATE requires less computation when representing one segment, thus can speed up the inference.

\subsection{Seg-Rep. via Segment Embeddings}\label{sec:seg-rep}

For segmentation problems, a segment is generally considered more informative and less ambiguous than an individual input.
Incorporating segment-level features usually lead performance improvement in previous semi-CRF work.
Segment representations in Section \ref{sec:alt-inp-rep} only model the composition of input units.
It can be expected that the segment embedding which encodes an entire subsequence as a vector can be an effective way for representing a segment.

\begin{figure}[t]
\centering
\includegraphics[width=0.7\columnwidth,trim={2cm 2.8cm 2cm 3.9cm},clip]{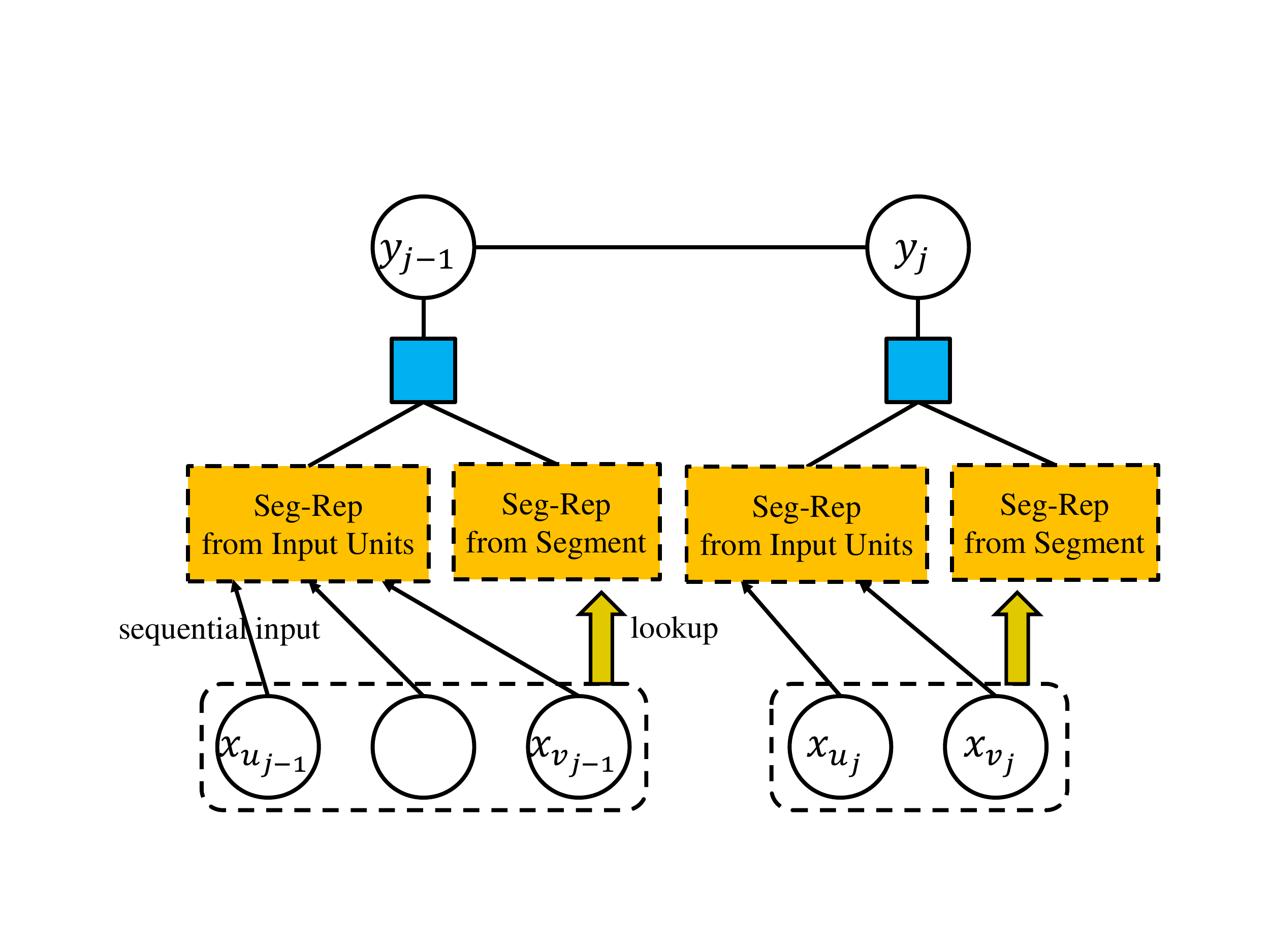}
\caption{Our neural semi-CRF model with segment representation from input composition and segment embeddings.}\label{fig:with-seg}
\end{figure}

In this paper, we treat the segment embedding as a lookup-based representation, which retrieves the embedding table with the surface string of entire segment.
With the entire segment properly embed, it is straightforward to combine the segment embedding with the composed vector from the input so that multi-level information of a segment is used in our model (see Figure \ref{fig:with-seg}).
However, how to obtain such embeddings is not a trivial problem.

A natural solution for obtaining the segment embeddings can be collecting all the ``correct'' segments from training data into a lexicon and learning their embeddings as model parameters.
However, the in-lexicon segment is a strong clue for a subsequence being a correct segment, which makes our model vulnerable to overfitting.
Unsupervised pre-training has been proved an effective technique for improving the robustness of neural network model \cite{Erhan:2010:WUP:1756006.1756025}.
To mitigate the overfitting problem, we initialize our segment embeddings with the pre-trained one.
 
Word embedding gains a lot of research interest in recent years \cite{DBLP:journals/corr/MikolovSCCD13} and is mainly carried on English texts which are naturally segmented.
Different from the word embedding works, our segment embedding requires large-scale segmented data, which cannot be directly obtained.
Following \textcite{wang-EtAl:2011:IJCNLP-2011} which utilize automatically segmented data to enhance their model, we obtain the auto-segmented data with our neural semi-CRF baselines (SRNN, SCNN, and SCONCATE) and use the auto-segmented data to learn our segment embeddings.

Another line of research shows that machine learning algorithms can be boosted by ensembling {\it heterogeneous} models.
Our neural semi-CRF model can take knowledge from heterogeneous models by using the segment embeddings learned on the data segmented by the heterogeneous models.
In this paper, we also obtain the auto-segmented data from a conventional CRF model which utilizes hand-crafted sparse features.
Once obtaining the auto-segmented data, we learn the segment embeddings in the same with word embeddings.

\begin{figure}[t]
\centering
\includegraphics[width=0.8\columnwidth,trim={0cm 0cm 3.5cm 11.2cm},clip]{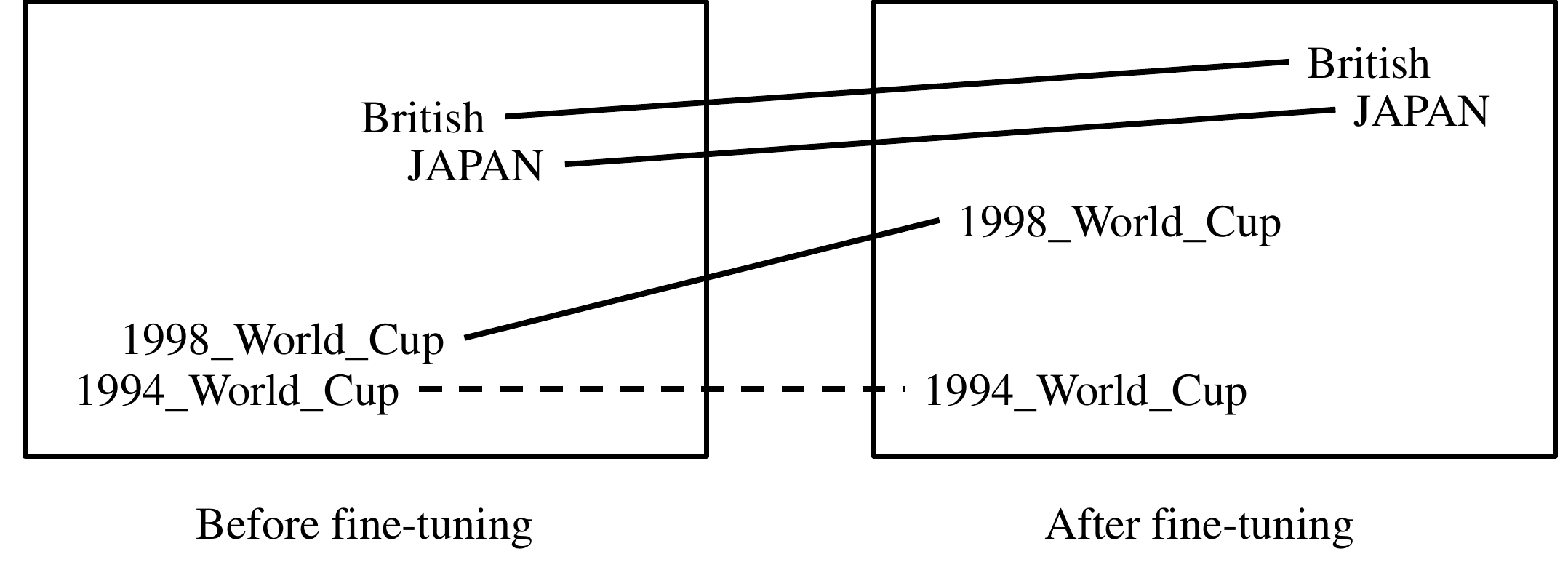}
\caption{An example for fine-tuning decreases the generalization power of pre-trained segment embedding.
``1994\_World\_Cup'' does not occur in the training data and its similarity with ``1998\_World\_Cup'' is broken because ``1998\_World\_Cup'' is tuned.}\label{fig:wo-ft}
\end{figure}

A problem that arises is the fine-tuning of segment embeddings.
Fine-tuning can learn a task-specific segment embeddings for the segments that occur in the training data, but it breaks their relations with the un-tuned out-of-vocabulary segments.
Figure \ref{fig:wo-ft} illustrates this problem.
Since OOV segments can affect the testing performance, we also try learning our model without fine-tuning the segment embeddings.

\subsection{Model details}

In this section, we describe the detailed architecture for our neural semi-CRF model.

\subsubsection{Input Unit Representation}
Following \textcite{DBLP:journals/corr/KongDS15}, we use a bi-LSTM to represent the input sequence.
To obtain the input unit representation, we use the technique in \textcite{dyer-EtAl:2015:ACL-IJCNLP} and separately use two parts of input unit embeddings: the pre-trained embeddings $E^p$ without fine-tuning and fine-tuned embeddings $E^t$.
For the $i$th input, $E^p_i$ and $E^t_i$ are merged together through linear combination and form the input unit representation 
\[I_i=\textsc{ReLU}(W^\mathcal{I} [E_i^p ; E_i^t] + b^\mathcal{I})\]
where the notation of $W[X_1;..;X_n]$ equals to $X_1,..,X_n$'s linear combination $W_1 X_1 + .. + W_n X_n$ and $b^I$ is the bias.
After obtaining the representation for each input unit, a sequence $(I_1,...,I_{|\mathbf{x}|})$ is fed to a bi-LSTM.
The hidden layer of forward LSTM $\overrightarrow{H_i}$ and backward LSTM $\overleftarrow{H_i}$ are combined as \[H_i=\textsc{ReLU}(W^\mathcal{H}[\overrightarrow{H_i};\overleftarrow{H_i}]+b^\mathcal{H})\]
and used as the $i$\textsuperscript{th} input unit's final representation.

\subsubsection{Segment Representation}
Given a segment $s_j=(u_j,v_j,y_j)$, a generic function \textsc{SComp}$(H_{u_j},...,H_{v_j})$ stands for the segment representation that composes the input unit representations $(H_{u_j},..,H_{v_j})$.
In this work, \textsc{SComp} is instantiated with three different functions: SRNN, SCNN and SCONCATE.
Besides composing input units, we also employ the segment embeddings as segment-level representation.
Embedding of the segment $s_j=(u_j,v_j,y_j)$ is denoted as a generic function \textsc{SEmb}$(x_{u_j}...x_{v_j})$ which converts the subsequence $(x_{u_j},...,x_{v_j})$ into its embedding through a lookup table.
At last, the representation of segment $s_j$ is calculated as
\[S_j=\textsc{ReLU}(W^\mathcal{S}[\textsc{SComp}_j;\textsc{SEmb}_j;E^Y_{y_j}]+b^\mathcal{S})\]
where $E^Y$ is the embedding for the label of a segment.

\begin{table}[t]
\small
\centering
\begin{tabular}{r|l}
\hline
fixed input unit embedding $E^p_i$ size      & 100 \\
fine tuned input unit embedding $E^t_i$ size & 32 \\
input unit representation $I_i$ size       & 100 \\
LSTM hidden layer $H_i$ size            & 100 \\
seg-rep via input composition {\sc SComp} & 64 \\
seg-rep via segment embedding {\sc SEmb} & 50 \\
label embedding $E^Y_{y_i}$ size & 20 \\
final segment representation $S_i$ size & 100 \\
\hline
\end{tabular}
\caption{Hyper-parameter settings}
\label{tbl:hyper-parameter}
\end{table}

Throughout this paper, we use the same hyper-parameters for different experiments as listed in Table \ref{tbl:hyper-parameter}.
\subsubsection{Training Procedure}

In this paper, negative log-likelihood is used as learning objective.
We follow \textcite{dyer-EtAl:2015:ACL-IJCNLP} and use stochastic gradient descent to optimize model parameters.
Initial learning rate is set as $\eta_0=0.1$ and updated as $\eta_t=\eta_0/(1+0.1t)$ on each epoch $t$.
Best training iteration is determined by the evaluation score on development data.

\section{Experiment}
We conduct our experiments on two NLP segmentation tasks: named entity recognition and Chinese word segmentation.

\subsection{Dataset and Word Embedding}

For NER, we use the CoNLL03 dataset which is widely adopted for evaluating NER models' performance.
F-score is used as evaluation metric.\footnote{{\tt conlleval} script in CoNLL03 shared task is used.}

For CWS, we follow previous study and use three Simplified Chinese datasets: PKU and MSR from 2\textsuperscript{nd} SIGHAN bakeoff and Chinese Treebank 6.0 (CTB6).
For the PKU and MSR datasets, last 10\% of the training data are used as development data as \cite{pei-ge-chang:2014:P14-1} does.
For CTB6 data, recommended data split is used.
We convert all the double byte digits and letters in the PKU data into single byte.
Like NER, CWS performance is evaluated by F-score.\footnote{{\tt score} script in 2\textsuperscript{nd} SIGHAN bakeoff is used.}

Unlabeled data are used to learn both the input unit embeddings (word embedding for NER, character embedding for CWS) and segment embeddings.
For NER, we use RCV1 data as our unlabeled English data.
For CWS, Chinese gigawords is used as unlabeled Chinese data.
Throughout this paper, we use the word embedding toolkit released by \textcite{ling-EtAl:2015:NAACL-HLT} to obtain both the input unit embeddings and segment embeddings.\footnote{\tt https://github.com/wlin12/wang2vec}

\subsection{Baseline}

We compare our models with three baselines:
\begin{enumerate}
\item {\sc Sparse-CRF}: The CRF model using sparse hand-crafted features.
\item {\sc NN-Labeler}: The neural network sequence labeling model making classification on each input unit.
\item {\sc NN-CRF}: The neural network CRF which models the conditional probability of a label sequence over the input sequence.
\end{enumerate}

BIESO-tag schema is used in all the CRF and sequence labeling models.\footnote{O tag which means OUTSIDE is not adopted in CWS experiments since CWS doesn't involve assigning tags to segments.}
For {\sc Sparse-CRF}, we use the baseline feature templates in \textcite{guo-EtAl:2014:EMNLP2014} for NER and 
\textcite{jiang-EtAl:2013:ACL2013}'s feature templates for CWS.
Both {\sc NN-Labeler} and {\sc NN-CRF} take the same input unit representation as our neural semi-CRF models but vary on the output structure and do not explicitly model segment-level information.

\subsection{Comparing Different Input Composition Functions}

\begin{table*}[t]
\small
\centering
\begin{tabular}{r|r||cc | cc cc cc| c}
\hline
 \multicolumn{2}{r||}{} & \multicolumn{2}{c|}{NER}  & \multicolumn{6}{c|}{CWS} & \\
 \multicolumn{2}{r||}{} & \multicolumn{2}{c|}{CoNLL03} & \multicolumn{2}{c}{CTB6} & \multicolumn{2}{c}{PKU} & \multicolumn{2}{c|}{MSR} & \\
 \multicolumn{2}{r||}{\it model} & dev & test & dev & test & dev & test & dev & test & spd \\
\hline
\multirow{3}{*}{\it baseline} & \sc NN-Labeler & 93.03 & 88.62 & 93.70 & 93.06 & 93.57 & 92.99 & 93.22 & 93.79 & \bf 3.30 \\
& \sc NN-CRF &\bf 93.06 &\bf 89.08 & 94.33 & 93.65 & 94.09 & 93.28 & 93.81 & 94.17 & 2.72 \\
\cline{2-11}
& \sc Sparse-CRF & 88.87 & 83.43 &\bf 95.68 &\bf 95.08 &\bf 95.85 &\bf 95.06 &\bf 96.09 &\bf 96.54 & \\
\hline
\multirow{3}{*}{\it neural semi-CRF}& \sc SRNN & 92.97 & 88.63 & 94.56 & 94.06 & 94.86 & 93.91 & 94.38 & 95.21 & 0.62 \\
& \sc SCONCATE & 92.96 & 89.07 & 94.34 & 93.96 & 94.41 & 93.57 & 94.05 & 94.53 & 1.08 \\
& \sc SCNN & 91.53 & 87.68 & 87.82 & 87.51 & 79.64 & 80.75 & 85.04 & 85.79 & 1.46 \\
\hline
\end{tabular}
\caption{The NER and CWS results of the baseline models and our neural semi-CRF models with different input composition functions.
{\it spd} represents the inference speed and is evaluated by the number of tokens processed per millisecond.}
\label{tbl:close-result}
\end{table*}

We first consider the problem of representing segments via composing input units and compare different input composition functions.
Results on NER and CWS data are shown in Table \ref{tbl:close-result}.
From this table, the SRNN and SCONCATE achieve comparable results and perform better than the SCNN.
Although CNN can model input sequence at any length, its invariance to the exact position can be a flaw in representing segments.
The experimental results confirm that and show the importance of properly handling the input position.
Considering SCNN's relatively poor performance, we only study SRNN and SCONCATE in the following experiments.

Comparing with {\sc NN-Labeler}, structure prediction models (NN-CRF and neural semi-CRF) generally achieve better performance.
The best structure prediction model outperforms {\sc NN-Labeler} by 0.4\% on NER and 1.11\% averagely on CWS according to Table \ref{tbl:close-result}.
But the difference between the neural structure prediction models is not significant.
NN-CRF performs better than the best neural semi-CRF model (SCONCATE) on NER while the both SRNN and SCONCATE outperform NN-CRF on three CWS datasets.
We address this to the fact either the NN-CRF or the neural semi-CRF merely takes input-level information and not sufficiently adopts segment-level information into the models.

A further comparison on inference speed shows that SCONCATE runs 1.7 times faster than SRNN, but slower than the {\sc NN-Labeler} and NN-CRF, which is resulted from the intrinsic difference in time complexity.

\subsection{Comparing Different Segment Embeddings}

Next we study the effect of different segment embeddings.
Using a segmentation model, we can obtain auto-segmented unlabeled data, then learn the segment embeddings.
In this paper, we tried two segmentation models.
One is the neural semi-CRF baseline which represents segment by composing input and another one is the CRF model using sparse hand-crafted features.
For convenience, we use {\sc SEmb-Homo} and {\sc SEmb-Hetero} to note the segment embeddings learned from their auto-segmented data respectively.

\subsubsection{Effect of Pre-trained Segment Embeddings}

\begin{figure}[t]
\centering
\includegraphics[width=\linewidth,trim={0cm 0.4cm 0cm 0.5cm},clip]{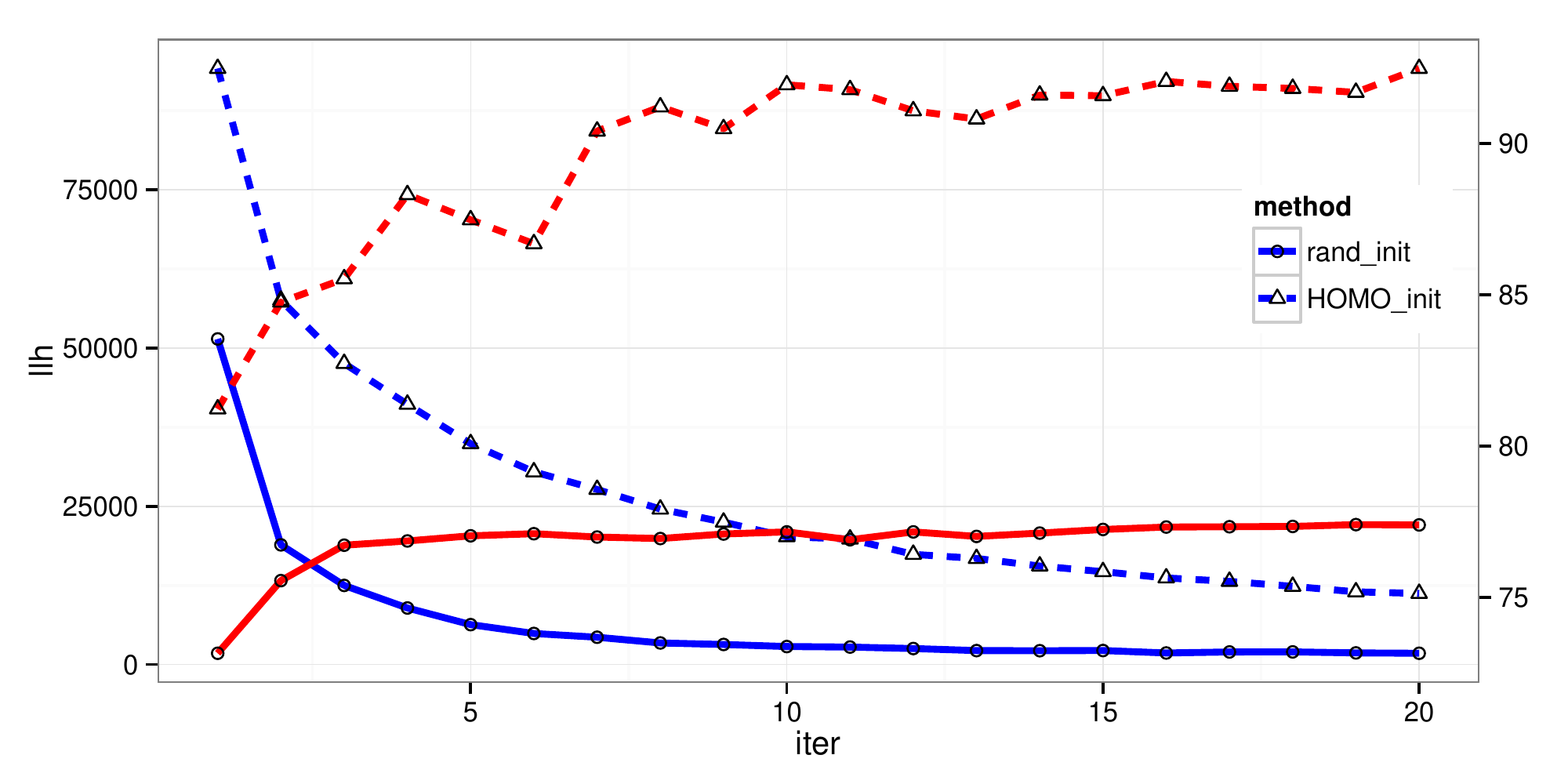}
\caption{Negative log-likelihood (blue lines) and development F-score (red lines) by iterations.
Solid lines show the model with randomly initialized segment embeddings.
Dashed lines show that initialized with pre-trained.}
\label{fig:learning-curve}
\end{figure}

We first incorporate randomly initialized segment embeddings into our model and tune the embeddings along with other parameters.
However our preliminary experiments of adding these embeddings into SRNN witness a severe drop of F-score on the CoNLL03 development set (from 92.97\% to 77.5\%).
A further investigation shows that the randomly initialized segment embeddings lead to severe overfitting.
Figure \ref{fig:learning-curve} shows the learning curve in training the NER model.
From this figure, the model with randomly initialized segment embeddings converge to the training data at about 5\textsuperscript{th} iteration and the development performance stops increasing at the same time.
However, by initializing with {\sc SEmb-Homo}, the development set performance increase to 93\%, which shows the necessity of pre-trained segment embeddings.

\subsubsection{Effect of Fine-tuning Segment Embeddings}

\begin{table}[t]
\small
\centering
\setlength{\tabcolsep}{5.5pt}
\begin{tabular}{r||c|ccc}
\hline
\it  model & CoNLL03 & CTB6 & PKU & MSR \\
\hline
  SRNN & 92.97 & 94.56 & 94.86 & 94.80 \\
\hdashline[1pt/3pt]
 \sc +SEmb-Homo w/FT & 92.97 & 95.83 &\bf 96.70 &\bf 97.32 \\
 \sc +SEmb-Homo wo/FT &\bf 93.14 &\bf 95.91 & 96.64 & 96.59 \\
\hline
 SCONCATE & 92.96 & 94.34 & 94.41 & 94.05 \\
\hdashline[1pt/3pt]
 \sc +SEmb-Homo w/FT & 93.07 & 95.79 &\bf 96.75 &\bf 97.29 \\
 \sc +SEmb-Homo wo/FT &\bf 93.36 &\bf 95.88 & 96.50 & 96.44 \\
\hline
 OOV & 46.02 & 5.45 & 5.80 & 2.60 \\
\hline
\end{tabular}
\caption{Effect of fine-tuning (FT) segment embedding on development data.
For CoNLL03 data, a named entity is ``out-of-vocabulary'' when it is not included in the training data as a named entity.}
\label{tbl:tuning-effect}
\end{table}

We study the effect of fine-tuning the segment embeddings by imposing {\sc SEmb-Homo} into our model.
Table \ref{tbl:tuning-effect} shows the experimental results on development data.
We find that our models benefit from fixing the segment embeddings on CoNLL03.
While on MSR, fine-tuning the embeddings helps.
Further study on the out-of-vocabulary rate shows that the OOV rate of MSR is very low, thus fine-tuning on segment embeddings help to learn a better task-specified segment representation.
However, on CoNLL03 data whose OOV rate is high, fine-tuning the segment embedding harms the generalization power of pre-trained segment embeddings.

\subsubsection{Effect of Heterogeneous Segment Embeddings}

\begin{figure}[t]
\includegraphics[width=\linewidth,trim={0cm 0.8cm 0cm 0.5cm},clip]{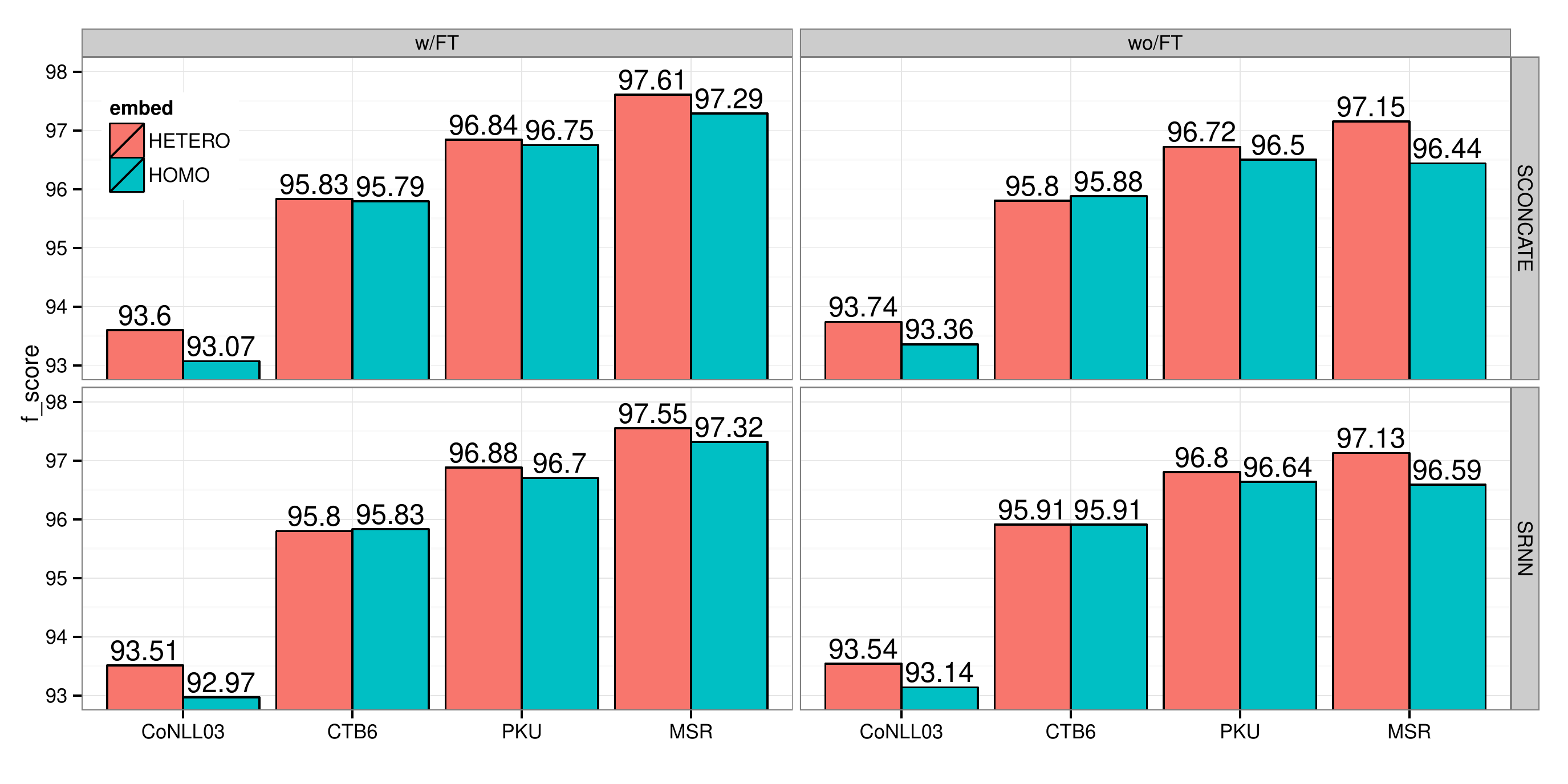}
\caption{Comparison between models with {\sc SEmb-Homo} and {\sc SEmb-Hetero} on development data.
The rows show different baseline neural semi-CRF models and the columns show whether fine-tuning (FT) the segment embeddings.}
\label{fig:bl-vs-crf}
\end{figure}

In previous sections, our experiments are mainly carried on the segment embeddings obtained from homogeneous models.
In this section, we use our {\sc Sparse-CRF} as the heterogeneous model to obtain {\sc SEmb-Hetero}.
We compare the models with {\sc SEmb-Hetero} and {\sc SEmb-Homo} on the development data in Figure \ref{fig:bl-vs-crf}.
These results show that {\sc SEmb-Hetero} generally achieve better performance than the {\sc SEmb-Homo}.
On the CoNLL03 and MSR dataset, the differences are significant.
Meanwhile, we see that fine-tuning the segment embedding can narrow the gap between {\sc SEmb-Hetero} and {\sc SEmb-Homo}.

\subsubsection{Final Result}

\begin{table}[t]
\small
\centering
\begin{tabular}{r||c |c c c}
\hline
\it  model & CoNLL03 & CTB6 & PKU & MSR \\
\hline
\sc NN-labeler & 88.62 & 93.06 & 92.99 & 93.79 \\
\sc NN-CRF & 89.08 & 93.65 & 93.28 & 94.17 \\
\hline
\sc Sparse-CRF & 83.43 & 95.08 & 95.06 & 96.54 \\
\hline
 SRNN &88.63 & 94.06 & 93.91 & 95.21 \\
\sc +SEmb-Hetero & 89.59 &\bf 95.48 & 95.60 & 97.39 \\
 &\it +0.96 &\it +1.42 &\it +1.69 &\it +2.18 \\
\hdashline[1pt/3pt]
 SCONCATE & 89.07 & 93.96 & 93.57 & 94.53 \\
\sc +SEmb-Hetero & \bf 89.77 & 95.42 &\bf 95.67 &\bf 97.58 \\
&\it +0.70 &\it +1.43 &\it +2.10 &\it +3.05 \\
\hline
\end{tabular}
\caption{Comparison between baselines and our neural semi-CRF model with segment embeddings.}
\label{tbl:seg-result}
\end{table}

At last, we compare our neural semi-CRF model leveraging additional segment embeddings with those only represent segment by composing input.
Table \ref{tbl:seg-result} shows the result on the NER and CWS test data.
Style of segment embeddings ({\sc Homo} or {\sc Hetero}) and whether fine-tune it is decided by the development data.
From this result, we see that segment-level representation greatly boost up model's performance.
On NER, an improvement of 0.7\% is observed and that improvement on CWS is more than 2.0\% on average.

We compare our neural semi-CRF model leveraging multi-levels segment representation with other state-of-the-art NER and CWS systems.
Table \ref{tbl:ne-stoa} shows the NER comparison results.
\begin{table}[t]
\small
\centering
\begin{tabular}{r|r||c}
\hline
\it genre & \it model & CoNLL03 \\
\hline
\multirow{2}{*}{\it NN} & \cite{Collobert:2011:NLP:1953048.2078186} & 89.59 \\
& \cite{DBLP:journals/corr/HuangXY15} & 90.10 \\
\hdashline[1pt/3pt]
\multirow{3}{*}{\it non-NN} & \cite{Ando:2005:FLP:1046920.1194905} & 89.31 \\
& \cite{guo-EtAl:2014:EMNLP2014} & 88.58 \\
& \cite{passos-kumar-mccallum:2014:W14-16} & \bf 90.90 \\
\hline
\multicolumn{2}{r||}{our best} & 89.77 \\
\hline
\end{tabular}
\caption{Comparison with the state-of-the-art NER systems.}\label{tbl:ne-stoa}
\end{table}
The first block shows the results of neural NER models and the second one shows the non-neural models.
All these work employed hand-crafted features like capitalization.
\textcite{Collobert:2011:NLP:1953048.2078186}, \textcite{guo-EtAl:2014:EMNLP2014}, and \textcite{passos-kumar-mccallum:2014:W14-16} also utilize lexicon as an additional knowledge resource.
Without any hand-crafted features, our model can achieve comparable performance with the models utilizing domain-specific features.

\begin{table}[t]
\small
\centering
\setlength{\tabcolsep}{5.4pt}
\begin{tabular}{r|r||c c c}
\hline
\it genre & \it model & CTB6 & PKU & MSR \\
\hline
\multirow{4}{*}{\it NN} & \cite{zheng-chen-xu:2013:EMNLP} & - & 92.4 & 93.3 \\
& \cite{pei-ge-chang:2014:P14-1} & & 94.0 & 94.9 \\
& \cite{pei-ge-chang:2014:P14-1} w/bigram & - & 95.2 & 97.2 \\
& \cite{DBLP:journals/corr/KongDS15} & & 90.6 & 90.7 \\
\hdashline[1pt/3pt]
\multirow{4}{*}{\it non-NN} & \cite{Tseng05aconditional} & - & 95.0 & 96.4 \\
& \cite{zhang-clark:2007:ACLMain} & - & 95.1 & 97.2 \\
& \cite{sun-EtAl:2009:NAACLHLT09} & - & 95.2 & 97.3 \\
& \cite{wang-EtAl:2011:IJCNLP-2011} &\bf 95.7 & - & - \\
\hline
\multicolumn{2}{r||}{our best} & 95.48 &\bf 95.67 &\bf 97.58 \\
\hline
\end{tabular}
\caption{Comparison with the state-of-the-art CWS systems.}\label{tbl:cws-stoa}
\end{table}

Table \ref{tbl:cws-stoa} shows the comparison with the state-of-the-art CWS systems.
The first block of Table \ref{tbl:cws-stoa} shows the neural CWS models and second block shows the non-neural models.
Our neural semi-CRF model with multi-level segment representation achieves the state-of-the-art performance on PKU and MSR data.
On CTB6 data, our model's performance is also close to \textcite{wang-EtAl:2011:IJCNLP-2011} which uses semi-supervised features extracted auto-segmented unlabeled data.
According to \textcite{pei-ge-chang:2014:P14-1}, significant improvements can be achieved by replacing character embeddings with character-bigram embeddings.
However we didn't employ this trick considering the unification of our model.

\section{Related Work}
Semi-CRF has been successfully used in many NLP tasks like information extraction \cite{NIPS2005_427}, opinion extraction \cite{yang-cardie:2012:EMNLP-CoNLL} and Chinese word segmentation \cite{andrew:2006:EMNLP,sun-EtAl:2009:NAACLHLT09}.
Its combination with neural network is relatively less studied.
To the best of our knowledge, our work is the first one that achieves state-of-the-art performance with neural semi-CRF model.

Domain specific knowledge like capitalization has been proved effective in named entity recognition \cite{ratinov-roth:2009:CoNLL}.
Segment-level abstraction like whether the segment matches a lexicon entry also leads performance improvement \cite{Collobert:2011:NLP:1953048.2078186}.
To keep the simplicity of our model, we didn't employ such features in our NER experiments.
But our model can easily take these features and it is hopeful the NER performance can be further improved.

Utilizing auto-segmented data to enhance Chinese word segmentation has been studied in \textcite{wang-EtAl:2011:IJCNLP-2011}.
However, only statistics features counted on the auto-segmented data was introduced to help to determine segment boundary and the entire segment was not considered in their work.
Our model explicitly uses the entire segment.

\section{Conclusion}

In this paper, we systematically study the problem of representing a segment in neural semi-CRF model.
We propose a concatenation alternative for representing segment by composing input units which is equally accurate but runs faster than SRNN.
We also propose an effective way of incorporating segment embeddings as segment-level representation and it significantly improves the performance.
Experiments on named entity recognition and Chinese word segmentation show that the neural semi-CRF benefits from rich segment representation and achieves state-of-the-art performance.

\section*{Acknowledgments}
This work was supported by the National Key Basic Research Program of China via grant 2014CB340503 and the National Natural Science Foundation of China (NSFC) via grant 61133012 and 61370164.

\bibliographystyle{named}

\end{document}